\pdfoutput=1

\documentclass[11pt]{article}

\usepackage{acl}
\usepackage{graphicx}
\usepackage{amssymb}

\usepackage{listings}

\usepackage{textcomp}
\usepackage{times}
\usepackage{latexsym}
\usepackage{subcaption}
\usepackage[font=Large]{subcaption}
\usepackage{amsmath}
\usepackage{amssymb}
\usepackage{amsthm}
\usepackage{comment}

\usepackage{multicol}
\usepackage{enumitem}
\usepackage{threeparttable}

\usepackage[T1]{fontenc}

\usepackage{booktabs}
\usepackage{xcolor}
\usepackage[most]{tcolorbox}
\usepackage{fancyvrb}
\newcommand{\blfootnote}[1]{%
  \begingroup
  \renewcommand\thefootnote{}\footnote{#1}%
  \addtocounter{footnote}{-1}%
  \endgroup
}

\newtcolorbox{promptbox}[1][]{
  enhanced,
  breakable,
  colback=gray!5,
  colframe=gray!40,
  boxrule=0.4pt,
  arc=2pt,
  left=8pt, right=8pt, top=6pt, bottom=6pt,
  fonttitle=\bfseries\small,
  coltitle=black,
  colbacktitle=gray!15,
  attach boxed title to top left={xshift=8pt, yshift=-2pt},
  boxed title style={
    colback=gray!15,
    colframe=gray!40,
    boxrule=0.3pt,
    arc=1pt,
  },
  #1
}

\usepackage[utf8]{inputenc}

\usepackage{microtype}

\usepackage{inconsolata}
\usepackage{mathtools}

\usepackage{graphicx}
\usepackage{algorithm}
\usepackage{algorithmic}

\usepackage{hyperref}
\usepackage[arabic,english]{babel}

\usepackage[most]{tcolorbox}
\usepackage{placeins}

\usepackage{booktabs}
\usepackage{multirow}
\usepackage{tabularx}
\usepackage{array}
\usepackage{bm}
\usepackage{siunitx}
\newcommand{\ourmethod}{{\textsc{LPM}}}

\usepackage{eqparbox}

\graphicspath{{figures/}}
\title{Latent Personal Memory: Represent personal memory as dynamic soft prompts}
\author{
  \begin{tabular}[t]{@{}c@{}}\textbf{Debrup Das}\textsuperscript{1,*,\dag} \\ \footnotesize\texttt{debrupdas@umass.edu}\end{tabular}
  \hspace{5mm}
  \begin{tabular}[t]{@{}c@{}}\textbf{Avinash Amballa}\textsuperscript{2,*} \\ \footnotesize\texttt{a.amballa@samsung.com}\end{tabular}
  \hspace{5mm}
  \textbf{Yashas Malur Saidutta}\textsuperscript{2} \\ \\
   \textbf{Vijay Srinivasan}\textsuperscript{2}  \hspace{5mm}
  \textbf{Vivek Kulkarni}\textsuperscript{2} \hspace{5mm}  \textbf{Srinivas Chappidi}\textsuperscript{2} \\ \\
  \textsuperscript{1}\textbf{University of Massachusetts Amherst} \hspace{5mm}
  \textsuperscript{2}\textbf{Samsung Research America} \\
}

\definecolor{vijaycolor}{RGB}{150,75,0}

\begin{document}
\maketitle

\maketitle
\blfootnote{\textsuperscript{*}Equal contribution.}
\blfootnote{\textsuperscript{\dag}Work during internship at Samsung Research America, AI-Center-Mountain View.}

\begin{abstract}

Personalizing large language models (LLMs) requires encoding long-term, user-specific behavioral patterns in a way that is computationally efficient, scalable, and compatible with a frozen base model. We present Latent Personal Memory (LPM), a scalable framework that represents user-specific history as a compact, persistent matrix of $N$ latent slots, that are interpretable. A shared cross-attention projection network maps these slots into dynamic, input-conditioned soft prompts that are prepended to the input of a frozen LLM. We evaluate LPM on PersonaMem v1 and LoCOMO benchmarks across Qwen3-1.7B, 4B, and 8B backbones. Results demonstrate that LPM outperforms LoRA and Prompt Tuning by up to \textbf{8.8 \%} and \textbf{54.4\%} in overall accuracy respectively on PersonaMem v1, while reducing KV-cache usage by over \textbf{64x}. On LoCoMo, LPM matches LoRA's accuracy with \textbf{120x} fewer trainable parameters. We also show that LPM's efficiency grows with context length and outperforms full-context at 128K context length.

\end{abstract}

\begin{figure*}[th]
\centering
\includegraphics[width=0.85\linewidth]{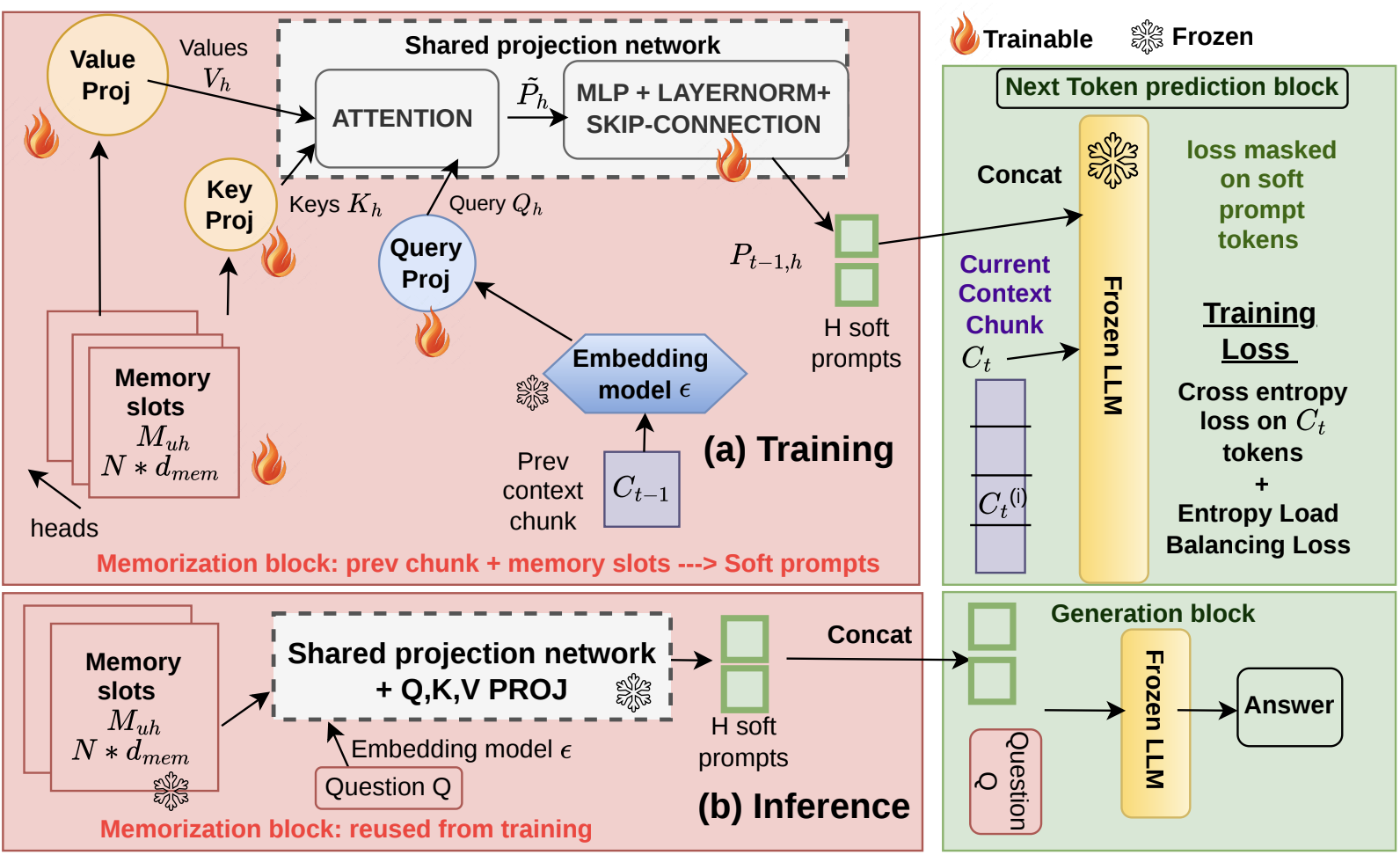}
\caption{\small (a) Training methodology of \ourmethod{} and (b) Inference setup for \ourmethod{}.}
\label{fig:lpm_design}
\end{figure*}
\section{Introduction \& Background}
\label{sec:introduction}

Personalizing large language models (LLMs) requires retaining long-term user-specific information while remaining \textit{computationally efficient}, scaling with a growing user base, minimizing storage, and preserving the model's general capabilities. Existing approaches struggle to jointly satisfy these requirements. \textbf{Full-context methods}~\cite{liu2025comprehensivesurveylongcontext} prepend the entire user history at inference, incurring \textit{high latency and large KV-cache overhead} due to the quadratic cost of self-attention. \textbf{RAG approaches}~\cite{lewis-rag, guo-etal-2025-lightrag, huang-rag, yuan-rag-selfreason} shorten prompts but maintain external stores that \textit{grow with user history}. Recent \textbf{text-based memory methods}~\cite{hu2026memoryageaiagents} store history as discrete structured units that are retrieved and updated as preferences evolve, but often rely on \textit{agentic architectures with large memory footprints} with multiple LLMs, hierarchical memories, and dedicated modules for memory updates~\cite{wang2025mirixmultiagentmemoryllmbased, wang2026memalpha, yan2026memoryr1enhancinglargelanguage}. Crucially, both RAG and text-based memory operate in the text space which is well-suited for exact factual recall, but \textit{poor at scaling with user history length and at abstracting how user preferences evolve}. \textbf{Parameter-efficient fine-tuning} methods like LoRA~\cite{hu2022lora} train small per-user weights but \textit{modify the base LLM, risking degradation of general reasoning}~\cite{shuttleworth2025lora}.  Prompt tuning~\cite{lester-etal-2021-prompttuning} learns static per-user soft prompts but suffers from \textit{limited expressivity}. Context-compression methods~\cite{huber2025embeddingtoprefixparameterefficientpersonalizationpretrained, tan-etal-2024-lloco} target generic compression rather than long-term personalization. \textbf{Latent memory methods} instead encode user information into the model's internal representations. Recent latent memory architectures with richer internal memory~\cite{wang2024memoryllmselfupdatablelargelanguage, wang2025mextendingmemoryllmscalable, memory3-yang, zhang2026memgen} require \textit{large-scale or continual pretraining} to learn memory read/write dynamics making them \textit{neither scalable nor interpretable}. See Appendix~\ref{sec:related_work} for further discussion on related works.

To overcome the above limitations, we introduce \textsc{Latent Personal Memory} (\textbf{LPM}), a scalable personalization framework that represents each user's history as a compact matrix of latent memory slots. A shared projection network dynamically retrieves question-conditioned soft prompts from these slots, which are prepended to the question and fed into the frozen LLM. Unlike static prompt tuning or LoRA, \textit{the retrieved prompts vary with the user query while keeping the backbone unchanged}. LPM offers four advantages: \textbf{(1) Scalability:} per-user memory is small and the projection network does not scale with users, \textbf{(2) Efficiency:} latent compression substantially reduces KV-cache, memory footprint, and latency versus full-context, \textbf{(3) Long context:} LPM's cost stays roughly constant as context grows while full-context inference scales sharply, \textbf{(4) Interpretability:} LPM offers interpretable memory in constrast to LoRA and prompt tuning. We evaluate LPM on PersonaMem v1~\cite{jiang2025knowmerespondme} and LoCoMo~\cite{maharana-etal-2024-evaluating} across Qwen3-1.7B, 4B, and 8B. On \textbf{PersonaMem v1}, LPM consistently outperforms LoRA and prompt tuning across all backbones, approaching full-context accuracy at up to \textbf{64 $\mathbf{x}$ smaller KV cache} and \textbf{4 $\mathbf{x}$ lower latency}. On \textbf{LoCoMo}, LPM matches LoRA's accuracy with $\sim$\textbf{120 $\mathbf{x}$  fewer trainable parameters}.

\section{Related Works}
\label{sec:related_work}

\paragraph{Full-context methods.} A direct approach to personalization is to prepend the user's entire interaction history to each query~\cite{liu2025comprehensivesurveylongcontext}. While conceptually simple, this incurs high latency and substantial computational overhead from large K-V cache sizes. The cost stems from the self-attention mechanism, whose complexity scales quadratically with sequence length, so inference costs grow rapidly with the number of input tokens.

\paragraph{Retrieval-augmented generation (RAG).} RAG approaches~\cite{lewis-rag, guo-etal-2025-lightrag, huang-rag, yuan-rag-selfreason} reduce prompt length by retrieving only relevant snippets from an external store. However, they rely on per-user indices that grow with history length and incur additional overhead for updates as new interactions accumulate.

\paragraph{Text-based memory methods.} A growing line of work~\cite{hu2026memoryageaiagents} stores user history as discrete units in structured form, retrieving, reorganizing, and updating entries as preferences evolve. Many of these adopt agentic architectures with high memory footprints, requiring multiple LLMs in memory, hierarchical memory levels, and dedicated modules for memory evolution and generation~\cite{wang2025mirixmultiagentmemoryllmbased, wang2026memalpha, yan2026memoryr1enhancinglargelanguage}. A smaller subset, including Mem0~\cite{chhikara2025mem0buildingproductionreadyai} and LightMem~\cite{fang2026lightmem}, instead targets scalable, efficient deployment with lower latency. All of these methods, however, leave the user memory \emph{external} to the LLM, maintaining a per-user index that grows with history length and preserves user history in the text space.

\paragraph{Parameter-efficient fine-tuning.} Methods such as LoRA~\cite{hu2022lora} train a small set of per-user weights efficiently. However, because they directly modify the base LLM's weights, they risk degrading its general reasoning capabilities as observed in some works~\cite{shuttleworth2025lora}.

\paragraph{Latent memory methods.} An alternative line of work encodes user information directly into the model's internal representations, offering a more compact alternative to text-based memory. Prompt tuning~\cite{lester-etal-2021-prompttuning} learns static per-user soft prompts concatenated before the question, compressing each user into a fixed, low-capacity representation that limits personalization expressivity. Context-compression methods~\cite{huber2025embeddingtoprefixparameterefficientpersonalizationpretrained, tan-etal-2024-lloco} train compressors that map long contexts into shorter soft-token sequences, but target generic compression rather than long-term personalization. More recent latent memory architectures introduce richer internal memory structures at higher cost: \textbf{MemoryLLM}~\cite{wang2024memoryllmselfupdatablelargelanguage} and \textbf{M+}~\cite{wang2025mextendingmemoryllmscalable} learn a self-updatable latent memory pool but require large-scale C4 / Common-Crawl-style pretraining to acquire their read/write dynamics. \textbf{Memory3}~\cite{memory3-yang} represents latent memory as K-V cache vectors written and read via a continual pretraining stage. \textbf{MemGen}~\cite{zhang2026memgen} generates latent memory tokens via a separate generator LLM, adding both parameter and GPU-memory costs.

\ourmethod{} also falls in the latent memory family but focuses \textit{specifically on efficient memory representation for personalization}. We build on a frozen pre-trained LLM, \textit{requiring no additional pretraining}, and pair a small per-user slot matrix with a shared projection network trained once across the user population. This yields a memory footprint that scales efficiently with the number of users.

\section{\ourmethod{} Methodology}
\label{sec:method}

LPM builds on the observation that parameter-efficient fine-tuning (PEFT) methods offer better accuracy and lower computational costs than few-shot In-Context Learning (ICL) \cite{liu2022fewshotparameterefficientfinetuningbetter}.
Rather than storing or re-processing raw conversation histories at inference time, Latent Personal Memory (\ourmethod{}) represents each user's history as a compact matrix of learned latent slots. A shared projection network then maps these slots into dynamic, input-conditioned soft prompts that are prepended to the input of the frozen LLM.

\textbf{Per-user latent slots.} Each user $u$ is associated with a persistent memory matrix $M_u \in \mathbb{R}^{N \mathrm{x} d_{\text{mem}}}$ of $N$ latent slots, that are learned during training (Fig \ref{fig:lpm_design}). These slots serve as a compressed, user-specific representation of the individual's conversational history. \footnote{In our Personamemv1 setting, there are only $\sim$65K unique parameters per user across all heads, independent of history length.}

\textbf{Shared projection network.} Given a context $x$, we compute embeddings $\epsilon(x)$ using a small frozen embedding model, then apply cross-attention between $\epsilon(x)$ and $M_u$ to obtain $\tilde{P}_h$. Here, the subscript $h$ represents the memory head.  
\begin{equation}
    \tilde{P_h} = \mathrm{softmax}\!\left(\tfrac{W_{Qh} e(x)\,(W_{Kh} M_{uh})^\top}{\sqrt{d}}\right) W_{Vh} M_{uh}.
\end{equation}
Sharing the projection network across users makes the approach scalable and compatible with federated learning, since user-specific slots can remain local while shared parameters are updated jointly. To avoid the bottleneck of a single memory matrix, we use $H$ independent memory heads $M_{uh}$, each with its own slots and projection. Per head, $\tilde{P}_h$ passes through an MLP and LayerNorm to produce $P_h(x, M_{uh}) \in \mathbb{R}^{d}$, where $d$ is the LLM hidden dimension. The equation for the MLP+LayerNorm pipeline used for computing the soft prompts of LPM is provided below. 

\begin{equation}
    P_h(x, M_{uh}) = \mathrm{LN_h}\!\big(\tilde{P_h} + \mathrm{MLP}(\tilde{P_h})\big).
\end{equation}

\noindent The $H$ prompts are concatenated and prepended to the input: \[\hat{y} = \mathrm{LLM}_{\text{frozen}}([P_{1}(x, M_{u1}); \ldots; P_{H}(x, M_{uH}); x])\] Because $x$ drives the attention query, the same memory $M_u$ yields different soft prompts for different inputs.

\begin{algorithm}[t]
\caption{\ourmethod{} training algorithm}
\label{alg:rotation}
\begin{algorithmic}[1]
\REQUIRE Users $\{u\}_{u=1}^{U}$ with contexts $H_u$; chunk size $S$; epochs $E$
\REQUIRE Shared projection $\theta$; per-user slots $\{M_u\}$
\FOR{$e = 1$ to $E$}
    \STATE Initialize accumulated gradient $g_\theta \leftarrow 0$
    \FOR{$u = 1$ to $U$}
        \STATE Load $M_u$; split $H_u$ into chunks $C_{1,u}, \ldots, C_{T_u, u}$
        \FOR{$t = 1$ to $T_u$}
            \IF{$t = 1$}
                \STATE $P \leftarrow P(C_{1,u}, M_u; \theta)$
            \ELSE
                \STATE $P \leftarrow P(C_{t-1, u}, M_u; \theta)$
            \ENDIF
            \STATE Compute $\mathcal{L}$ on chunk $C_{t,u}$ using soft prompt $P$
            \STATE Accumulate $g_\theta \leftarrow g_\theta + \nabla_\theta \mathcal{L}$
            \STATE Update $M_u$ by gradient step on $\nabla_{M_u} \mathcal{L}$
        \ENDFOR
        \STATE Save $M_u$
    \ENDFOR
    \STATE Update $\theta$ using accumulated gradient $g_\theta$
\ENDFOR
\ENSURE Trained shared projection $\theta$; per-user slots $\{M_u\}$
\end{algorithmic}
\end{algorithm}

\subsection{Training.}\label{sec:training} The user's full history $C_{user}$ is split into fixed-size chunks processed sequentially to compress context into memory. For each chunk $C_t$, the model performs next-token prediction \cite{tandon2025endtoendtesttimetraininglong} using a soft prompt retrieved from current memory slots, augmented with a hinge-style entropy regularizer motivated by mixture-of-experts works \cite{zoph2022stmoedesigningstabletransferable, li2025dynmoleboostingmixturelora}:
\begin{equation}
\label{eq:loss}
\begin{aligned}
    \mathcal{L} = &-\sum_{i} \log p\!\left(C_t^{(i)} \mid C_t^{(<i)},\, P_{t-1}\right) \\
    &+ \alpha \cdot \max\!\big(0,\, H_{\text{target}} - H(a)\big),
\end{aligned}
\end{equation}

where $H(a)$ is the entropy of the slot-attention distribution and $H_\mathrm{target}$ is an hyperparameter set to $\mathrm{0.4*log(k)}$ where $k$ is the number of memory slots in each head. The asymmetric regularizer penalizes attention collapse but never forces uniformity, preventing semantic redundancy across slots while allowing concentration when appropriate. The prompt $P_{t-1}$ is retrieved from the \emph{previous} chunk $C_{t-1}$  to prevent future-token leakage (Fig \ref{fig:lpm_design} (a)). We iterate over users and, per user, over context chunks, updating the active user's slots $M_{uh}$ at each step. Projection-network gradients are accumulated across all users and applied once per epoch to avoid noisy updates (Algorithm~\ref{alg:rotation}). Trainable parameters are $W_{Qh}, W_{Kh}, W_{Vh}$, the MLP, LayerNorm, and $\{M_{uh}\}$; the base LLM stays frozen. See Appendix~\ref{app:chat-template} for chat-template details.

\subsection{Inference.} At inference, LPM uses only the user's question to retrieve soft prompts, which are prepended to the question tokens and passed to the frozen LLM, yielding a short effective context and low latency (Fig \ref{fig:lpm_design} (b)). The cost of processing each user's history is paid once during training and amortized over all subsequent inference queries, making LPM significantly more efficient than full-context or RAG-based approaches at deployment.

\section{Experiments}
\label{sec:datasets}

\begin{table*}[t]
\centering
\caption{\textbf{PersonaMem v1 main results: Qwen3-8B.} Accuracy
(\%) across PersonaMem question categories, together with inference
efficiency. Means are
computed across all questions. Question types and notations in Table \ref{tab:personamem-stats}.}
\label{tab:personamem_8b}
\renewcommand{\arraystretch}{1.1}
\setlength{\tabcolsep}{3pt}
\resizebox{\linewidth}{!}{%
\begin{tabular}{l ccccccc c ccccc}
\toprule
& \multicolumn{7}{c}{\textbf{Accuracy (\%)} $\uparrow$}
& & \multicolumn{5}{c}{\textbf{Efficiency} $\downarrow$} \\
\cmidrule(lr){2-8} \cmidrule(lr){10-14}
\textbf{Method}
& \textbf{R-Fact} & \textbf{Sug} & \textbf{P-Evol} & \textbf{P-Rec}
& \textbf{P-Rsn} & \textbf{Gen} & \textbf{Overall}
& & \textbf{Extra} & \textbf{Ctx} & \textbf{KV (MB)}
& \textbf{Peak (MB)} & \textbf{Lat. (s)} \\
\midrule
Full-context  & \underline{46.60} & 6.50  & \textbf{73.40} & 41.80 & \textbf{81.80} & \textbf{59.60} & \textbf{53.30} & & 0      & 25943.2 & 3648.27 & 23001.9 & 3.81 \\
RAG (top-8)          &  \textbf{52.74}   & \underline{20.43} & 54.03 & 41.82   &  64.85  &  46.88   & 44.50 & & 0      & 2232.0  & 313.88  & 16177.5 & 1.79 \\
LightMem (top-50)     & 44.52    & \textbf{23.65}    & 35.25    &  \textbf{54.54}    & 54.54 & 45.61    & 41.77 & & 0      & 2523.8  & 354.77  & 16248.1 & 1.67 \\
LoRA          & 33.60 & 17.20 & 61.20 & 30.90 & 74.70 & 35.10 & 44.30 & & 154\,M & \textbf{403.5}   & \underline{56.73}   & \textbf{15671.4} & 5.44 \\
Prompt Tuning & 33.60 & 6.50  & 30.90 & 32.70 & 53.50 & 26.30 & 31.24 & & \underline{40\,M}  & 953.5   & 120.0   & 16302.5 & 1.65 \\
Embedding-to-Prefix & 28.08 & 9.68  & 59.71 &  34.55 & 70.71  &  31.58 & 40.75 & & 346\,M & 411.5  & 55.54   & 16942.1 &  \textbf{0.36} \\
\midrule
\textbf{\ourmethod{} (Ours)}
              & 35.62 & 3.23 & \underline{66.91}
              & \underline{45.45} & \underline{80.81} & \underline{54.39}
              & \underline{48.22}
              & & \textbf{1.3\,M} & \underline{411.5} & \textbf{56.73}
              & \underline{16216.8} & \underline{0.88} \\
\bottomrule
\end{tabular}%
}
\end{table*}
\subsection{Datasets}
We evaluate \ourmethod{} on two long-context personalization benchmarks: PersonaMem v1~\cite{jiang2025knowmerespondme}, a multiple-choice personalization benchmark, and LoCOMO~\cite{maharana-etal-2024-evaluating}, a generative long-context QA benchmark.
\paragraph{PersonaMem v1.}
PersonaMem v1~\cite{jiang2025knowmerespondme} is a long-range
personalization benchmark containing approximately 32K-token contexts
per user together with multiple-choice questions spanning six in-situ
personalization categories: factual recall, suggesting new ideas,
tracking preference evolution, reasons behind preference updates,
preference-aligned recommendations, and generalization to new
scenarios. We evaluate on all 20 users in the dataset and report multiple-choice accuracy.
The per-category question counts are summarized in
Table~\ref{tab:personamem-stats}.
 
\paragraph{LoCOMO.}
LoCOMO~\cite{maharana-etal-2024-evaluating} is a long-conversation
benchmark consisting of multiple dialogue sessions between two
individuals with average context lengths of approximately 9K tokens.
The benchmark contains single-hop, multi-hop, temporal, and
open-domain questions requiring long-context reasoning over
conversations. Following prior work, we use GPT-5-mini as a judge
model \cite{chhikara2025mem0buildingproductionreadyai} to evaluate free-form generations against gold answers using
binary CORRECT/WRONG scoring. The per-category question counts are
summarized in Table~\ref{tab:locomo-stats}.

Alongside task performance, we report \textbf{inference efficiency metrics}: \textit{context length}, \textit{KV-cache memory}, \textit{peak GPU memory}, \textit{latency}, and \textit{trainable parameters}.

\paragraph{Experimental Setup.}
We evaluate on Qwen3-1.7B, 4B, and 8B backbones. Considering realistic deployment, we assume access only to long user histories $C_{user}$ without user-specific QA supervision. Unlike \textit{context distillation} approaches \cite{Snell2022Learningcontext} that rely on synthetically generated supervision, \textbf{we avoid generating QA pairs from the context, as such methods introduce additional computational overhead and are difficult to scale}. Hyperparameters and compute infrastructure for both datasets are provided in Table~\ref{tab:hyperparameters}. We compare \ourmethod{} against several baselines: (i) \textbf{Full-context}, where the user's entire history is fed at inference time, (ii) \textbf{RAG}, where a retriever selects relevant context for each query, (iii) \textbf{LightMem} \cite{fang2026lightmem}, a lightweight and efficient text memory-augmentation baseline, (iv) \textbf{LoRA} \cite{hu2022lora}, training low rank adaption weights for each user and (v) \textbf{Prompt Tuning} \cite{lester-etal-2021-prompttuning}, per-user soft prompts with 450 prompt tokens, (vi) \textbf{Embedding-to-prefix} (8 prefix)  \cite{huber2025embeddingtoprefixparameterefficientpersonalizationpretrained}: learns a trainable projection to map context embeddings to prefixes. Above three methods are trained sequentially with next token prediction on the context chunks similar to LPM.

\section{Results and Discussion}
\label{sec:results}

\subsection{PersonaMem v1 (Qwen3-8B).}

Table~\ref{tab:personamem_8b} shows that \ourmethod{} substantially outperforms baselines LoRA (\textbf{8.8\%}), Prompt Tuning (\textbf{54.4\%}), LightMem (\textbf{15.4\%}), RAG (\textbf{8.3\%}) while approaching full-context performance at a fraction of the inference cost. The question type-wise accuracy shows \ourmethod{} achieves the strongest gains on \textbf{reasons behind preference updates}, \textbf{tracking preference evolution}, and \textbf{generalizing to new scenario} questions, which are closer to \textit{semantic personalization} than \textit{exact facts recall}. These results align with our intuition that \textit{latent memory slots are better suited for compact, abstract representations of user preferences than for more factual retrieval}. On the efficiency side, \ourmethod{} reduces KV-cache usage by over \textbf{64 times} (from 3.65 GB to 56.73 MB) while achieving low inference latency .

\subsection{Scaling across backbone sizes.}
Table~\ref{tab:scaling} shows results across Qwen3-1.7B, 4B backbones. 
\begin{table}[h]
    \centering
    \resizebox{0.8\linewidth}{!}{
    \begin{tabular}{lcc}
    \toprule
    \textbf{Method} & \textbf{Qwen3-1.7B} & \textbf{Qwen3-4B}  \\
    \midrule
    Full-context     & \textbf{36.84} & \textbf{48.30}  \\
    RAG (top-8)     & \underline{31.88} & 42.42  \\
    LightMem (top-50)    & 29.54 & 38.20 \\
    LoRA             & 22.92 & 39.70  \\
    Prompt Tuning    & 18.85 & 38.70  \\
    Embedding-to-Prefix & 21.90 & 43.46  \\
    \midrule
    \textbf{\ourmethod{} (Ours)} & 30.05 & \underline{44.70}  \\
    \bottomrule
    \end{tabular}
    }
    \caption{\small \textbf{Scaling across backbones}: PersonaMem v1 overall accuracy (\%).}
    \label{tab:scaling}
\end{table}
\ourmethod{} consistently outperforms LoRA, Prompt Tuning, and LightMem across all model sizes. The gains are especially pronounced for smaller models, where latent memory provides stronger personalization benefits under limited model capacity.

\subsection{LPM vs Full-context on the 128K-context split of PersonaMem v1.}
\begin{figure}[h]
    \centering
    \includegraphics[width=1\linewidth]{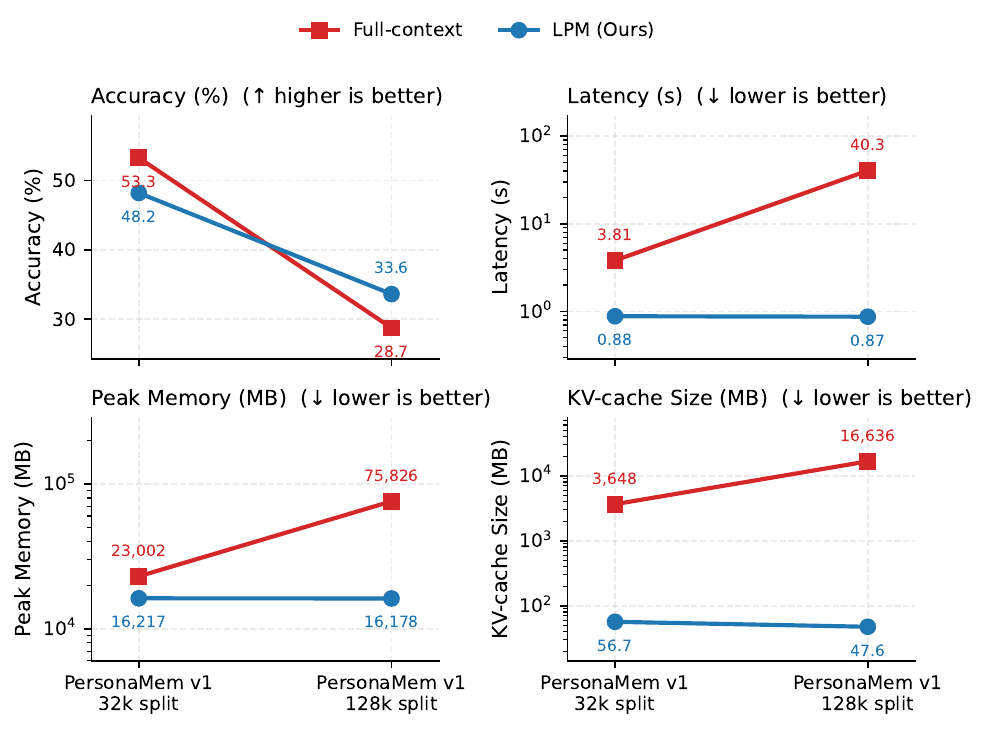}
    \caption{Scaling behavior of Full-context vs \ourmethod{} on
    PersonaMem v1 as context grows from ${\sim}32$K to
    ${\sim}128$K tokens.}
    \label{fig:personamem_scaling}
\end{figure}

We further evaluate on the long-context split of PersonaMem v1, where context lengths grow from $32$K to $128$K tokens. As shown in Figure~\ref{fig:personamem_scaling}, Full-context accuracy degrades below \ourmethod{} by $14.5 \%$ at 128K context length. 
At the same time, \ourmethod{}'s peak memory, latency and KV cache size stays essentially flat, whereas full-context grows with the input length.
%\yashas{We cannot say our method holds steady, we should say shows lesser decline than full context.}

\subsection{LoCOMO Results.}
Table~\ref{tab:locomo-8b} reports results on LoCOMO on Qwen3-8B. Full-context achieves the highest accuracy ($71.10\%$), likely because the average context length of 8k remains manageable for direct conditioning during generation. Among other approaches, \ourmethod{} achieves performance comparable to LoRA ($33.25\%$ vs.\ $33.10\%$) while requiring nearly $120\mathbf{x}$ fewer parameters ($0.65$M vs.\ $77$M). \ourmethod{} also improves inference efficiency by 3 times, shrinks KV-cache usage from $2.9$GB to only $15.3$MB compared to full context. Similar trends are observed across Qwen3-4B and Qwen3-1.7B as shown in Table~\ref{tab:locomo-other-backbones}. Additional ablations on LoCoMo are provided in Appendix \ref{app:ablations}.

\begin{table}[t]
\centering

\setlength{\tabcolsep}{2.6pt}
\renewcommand{\arraystretch}{0.95}
\caption{LoCOMO results on \textsc{Qwen3-8B}. Question-type columns correspond to Single-hop (SH), Multi-hop (MH), Temporal (Temp.) and Open-domain (Open.) accuracy. \textbf{The efficiency stats are mean values across all questions.} }
\label{tab:locomo-8b}
\resizebox{\linewidth}{!}{
\begin{tabular}{lcccccccccc}
\toprule
& \multicolumn{5}{c}{\textbf{Accuracy (\%)} $\uparrow$}
&  \multicolumn{5}{c}{\textbf{Efficiency} $\downarrow$} \\
\cmidrule(lr){2-6} \cmidrule(lr){7-11}
Method & MH & Open & Temp & SH. & Acc & Extra & Ctx & KV & Mem & Lat. \\
\midrule
Full-context  & 64.5 & 54.2 & 36.1 & 88.6 & 71.10 & 0 & 20626.0 & 2900.5 & 21370.3 & 3.43 \\
RAG  & 47.2 & 36.5 & 23.1 & 65.4 & 51.42 &0 & 4232.57& 595.20& 16663.80 & 2.85 \\
% LightMem  & -- & -- & -- & -- & &  & &  & & \\
LoRA          & 30.8 & 33.3 & 7.7 & 43.5 & 33.12 & 77M & 45.2 & 6.4 & 15676.9 & 3.53 \\
Prompt Tun.   & 31.9 & 29.2 & 12.8 & 39.4 & 31.88 & 20M & 495.2 & 69.6 & 15817.7 & 0.87 \\
\midrule
\textbf{\ourmethod{}} & 31.2 & 45.8 & 10.9 & 41.0 & \textbf{33.25} & \textbf{0.65M} & 45.2 & 15.3 & 18835.5 & 1.19 \\
\bottomrule
\end{tabular}
}
\end{table}

\begin{table}[h]
\centering

\setlength{\tabcolsep}{3pt}
\renewcommand{\arraystretch}{1.0}
\caption{LoCOMO results on \textsc{Qwen3-4B} and \textsc{Qwen3-1.7B}.
Question-type columns correspond to Single-hop (SH), Multi-hop (MH),
Temporal (Temp.) and Open-domain (Open.) accuracy.}
\label{tab:locomo-other-backbones}
\resizebox{\linewidth}{!}{%
\begin{tabular}{l ccccc c ccccc}
\toprule
& \multicolumn{5}{c}{\textsc{Qwen3-4B}} & & \multicolumn{5}{c}{\textsc{Qwen3-1.7B}} \\
\cmidrule(lr){2-6} \cmidrule(lr){8-12}
Method & MH & Open & Temp & SH & Acc & & MH & Open & Temp & SH & Acc \\
\midrule
Full-context          & 62.8 &  38.5 & 25.9 & 80.5 & 63.3 & & 53.9 & 48.9 & 35.2 & 81.2 & 64.6 \\
% RAG                   & -- & -- & -- & -- & -- & & -- & -- & -- & -- & -- \\
% LightMem              & -- & -- & -- & -- & -- & & -- & -- & -- & -- & -- \\
LoRA                  &  28.2 & 33.1 & 7.9 & 41.1 & 31.3 &  & 18.8 & 32.3  & 8.4 &  31.7 & 24.5 \\
Prompt Tun.           & 28.4 & 36.5 & 10.6 & 37.3 & 30.1 & & 18.8 & 29.1 & 9.0 & 32.8 & 25.0 \\
\textbf{\ourmethod{}} & 30.5 & 38.5 & 14.6 & 41.0 & 33.4 & & 22.3 & 36.5 & 8.7 & 32.9 & 26.2 \\
\bottomrule
\end{tabular}%
}
\end{table}

\subsection{General Capability preservation.}
Table \ref{tab:capability-preservation-small} shows that \ourmethod{} (trained on PersonaMem) largely preserves the base model's capabilities. \ourmethod{} matches the base model on GSM-8K and MMLU datasets (nine-subject split of MMLU)\\
\begin{table}[h]
    \centering
    \resizebox{0.6\linewidth}{!}{
    \begin{tabular}{lcc}
    \toprule
    \textbf{Method} & \textbf{GSM-8K} & \textbf{MMLU} \\
    \midrule
    Base model & \textbf{92.3} & 68.47 \\
    LoRA & 90.7 & 67.35 \\
    \textbf{Ours (LPM)} & 90.8 & \textbf{70.27} \\
    \bottomrule
    \end{tabular}
    }
    \caption{General reasoning preservation on \textsc{Qwen3-8B} with model trained on PersonaMemv1.}
    \label{tab:capability-preservation-small}
\end{table}

\subsection{Ablation Studies}
\label{app:ablations}

In this section we ablate two design choices of \ourmethod{}: (i) the
\textbf{entropy load balancing loss term} of LPM, and
(ii) \textbf{the number of memory heads}. For both ablations we keep all other
hyperparameters fixed and report per-category accuracy on the LOCOMO
benchmark together with the overall average. Our reference configuration
(\textbf{\ourmethod{}}) uses 64 memory heads and the asymmetric entropy
regularizer loss term described in Section~\ref{sec:method}.
\begin{table}[h]
\centering
\setlength{\tabcolsep}{4pt}
\resizebox{\linewidth}{!}{%
\begin{tabular}{llccccc}
\toprule
Ablation & Variant & SH & MH & Temp. & Open. & Acc. \\
\midrule
\multirow{2}{*}{Entropy term}
  & \ourmethod{} (default, w entropy)            & \textbf{41.0}  & \textbf{31.2}           & 10.9          & \textbf{45.8}           & \textbf{33.25} \\
  & \quad w/o entropy   & 36.15          & 28.37 & \textbf{14.33} & 38.54 & 30.32 \\
\midrule
\multirow{2}{*}{No of memory heads}
  & \ourmethod{} (default, 64 heads)  & \textbf{41.0} & \textbf{31.2}           & \textbf{10.9}          & \textbf{45.8}           & \textbf{33.25} \\
  & \quad 8 heads                     & 31.87          & 23.76          & 6.23           & 37.50 & 25.39 \\
\bottomrule
\end{tabular}%
}
\caption{Ablations of the entropy regularizer and the number of memory
heads on LOCOMO. The default \ourmethod{} configuration (64 heads,
w/ entropy) is repeated as the reference row for each ablation. All
other components are held fixed.}
\label{tab:ablations}
\end{table}

\begin{figure*}[t]
    \centering
    \includegraphics[width=0.5\linewidth]{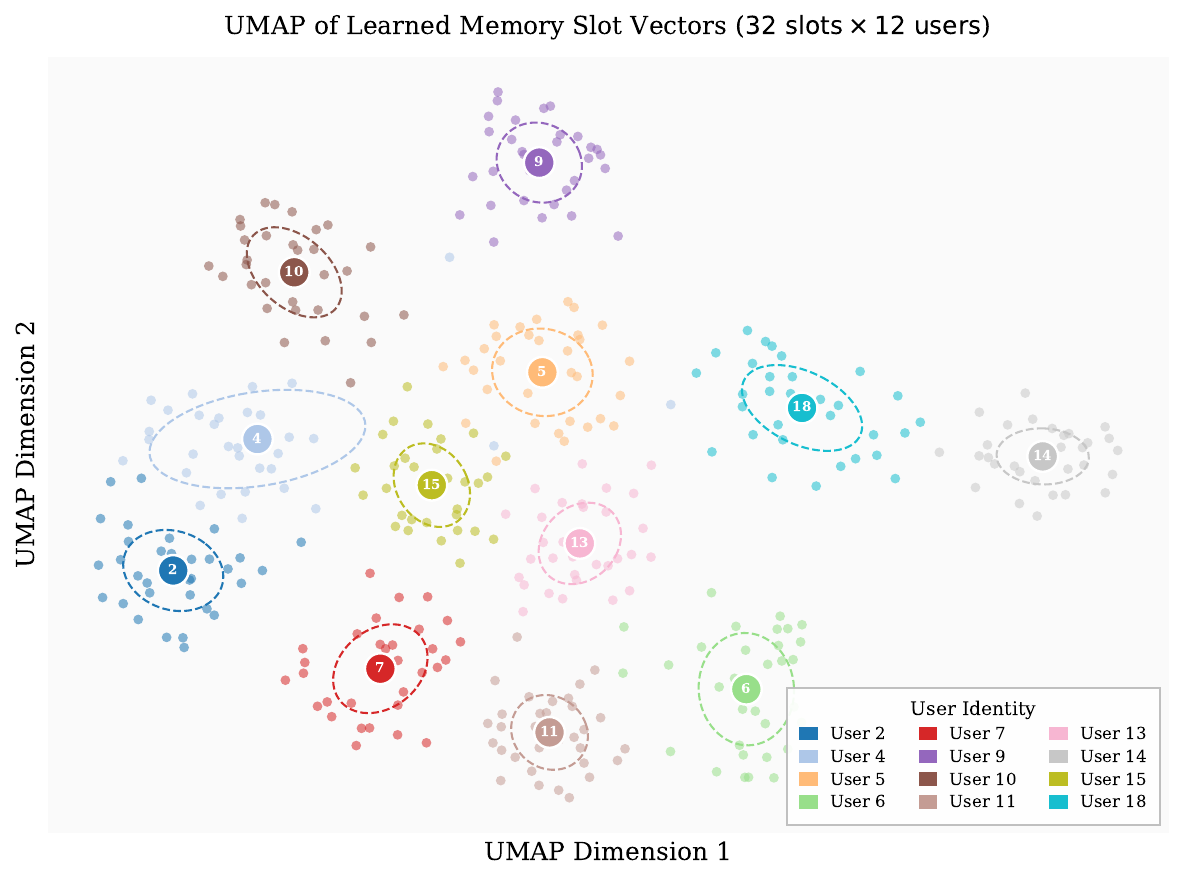}
    \caption{\small UMAP projection of the 32 learned memory slots for Personamem users. Each color denotes a distinct user.}
    \label{fig:umap_slots}
\end{figure*}
\subsubsection{Effect of the entropy regularization term.}
Table \ref{tab:ablations} shows that removing the entropy-based load balancing loss term leads to a drop in average accuracy (33.25 $\rightarrow$ 30.32). The default configuration improves substantially on \textbf{single-hop} (41.0 vs.\ 36.15), \textbf{multi-hop} (31.2 vs.\ 28.37), and \textbf{open-domain} (45.8 vs.\ 38.54), while removing the regularizer yields a modest gain on temporal questions (10.9 $\rightarrow$ 14.33). Since single-hop questions dominate LoCoMo (841 out of 1540 questions; see Table~\ref{tab:locomo-stats}), the single-hop gain is the largest contributor to the higher overall average. In practice, the regularizer additionally acts as a safeguard against head collapse, which we observed in early runs of training without it.

\subsubsection{Effect of the number of memory heads.}
Reducing the \textbf{number of heads from 64 to 8 degrades average accuracy substantially} (33.25 $\rightarrow$ 25.39) as shown in Table \ref{tab:ablations}. The drop is across all types but sharpest on \textbf{single-hop} (41.0 $\rightarrow$ 31.87) and \textbf{open-domain} (45.8 $\rightarrow$ 37.50), with notable drops on multi-hop (31.2 $\rightarrow$ 23.76) and temporal (10.9 $\rightarrow$ 6.23) as well. No category benefits from the smaller head count, indicating that the additional memory capacity from 64 heads is broadly useful across question types. We therefore adopt 64 heads as the default configuration.

\section{Analysis of Learned Memory Representations}
\label{sec:interpretability} 
% \vspace{-6mm}
\textit{A key advantage of our approach is that the trainable memory slots learn user-specific representations.} To verify this, we perform a geometric analysis on the PersonaMemv1. For each of the 20 users, we average the trained slot matrices across 8 heads to obtain $\mathbf{M}_u \in \mathbb{R}^{32\mathrm{x}256}$, then project all slot vectors into 2D with UMAP~\cite{mcinnes2020umapuniformmanifoldapproximation} (Figure~\ref{fig:umap_slots}). The slots form coherent, well-separated per-user clusters, suggesting they capture user-specific behavioral structure. To test whether this geometric separation reflects meaningful behavioral differences, we conduct a \textit{rubric-based qualitative similarity study} on representative users (2, 4, 14, 15). For each pair, the full conversational histories ($\sim$32K tokens) were given to \textsc{Claude-Opus-4.7} as judge, which scored pairwise similarity along dimensions including \textit{domain-specific preferences, communication style, personality traits} (rubric in Table~\ref{tab:rubric_dimensions}; full protocol in Appendix~\ref{app:user_similarity}). Table~\ref{tab:similarity_scores} reports representative scores. Notably, all pairings involving User~14 (the farthest from other users in UMAP space) receive the lowest similarity scores, aligning the geometric and rubric-based views. Example judge outputs for the most and least similar pairs appear in Tables~\ref{tab:most_similar_pair}and~\ref{tab:most_dissimilar_pair}.
Further qualitative findings are in Appendix~\ref{qual-findings}. Together, these results indicate that the learned memory slots are interpretable and encode behavioral traits beyond topical content.

\begin{table}[h]
\centering
\setlength{\tabcolsep}{8pt}
\resizebox{\linewidth}{!}{%
\begin{tabular}{lcc}
\toprule
User Pair & Mean Score & Qualitative Label \\
\midrule
User 2 $\leftrightarrow$ User 4  & 3.11 & Moderately--Highly Similar \\
User 2 $\leftrightarrow$ User 15 & 2.67 & Moderately Similar \\
User 4 $\leftrightarrow$ User 15 & 2.67 & Moderately Similar \\
User 4 $\leftrightarrow$ User 14 & 2.33 & Somewhat Similar \\
User 2 $\leftrightarrow$ User 14 & 2.22 & Somewhat Similar \\
User 15 $\leftrightarrow$ User 14 & 1.89 & Weakly Similar \\
\bottomrule
\end{tabular}
}
\caption{Representative rubric-based pairwise user similarity scores from LLM judge.}
\label{tab:similarity_scores}
\end{table}

\label{sec:conclusion}
\section{Conclusion}
In this paper. we introduce \ourmethod{}, a scalable framework for LLM personalization that represents each user's history as a small persistent matrix of latent slots and projects them into soft prompts via a shared projection network, making it scalable. Across PersonaMem and LoCOMO, \ourmethod{} consistently outperforms LoRA and Prompt Tuning with \textit{fewer parameters} and \textit{lower inference costs}. Geometric analysis of the learned memory slots reveals that \ourmethod{} learns \textit{interpretable latent representations} that capture personalized behavioral patterns.

\section*{Limitations}

Our work is not without its limitations. \\ \emph{(i) Hybrid approaches:}
LPM under-performs in memorizing factual knowledge. Future works can explore hybrid approaches combining latent and text-based memory approaches for personalization. A complementary explicit fact cache may be needed for applications that require exact recall.  

\emph{(ii) Memory granularities:} LPM relies on the fact that all heads equally capture long-term and short-term memory. Structuring memory heads to capture long-term vs.\ short-term memory of the user using heuristics, can be a future direction of research. 

\section{Acknowlegements.} We thank Haris Jeelani, Samsung Research America for help on baseline evaluations. This work was supported entirely by Samsung Research America, AI-Center Mountain View. 

\bibliography{custom}

@inproceedings{lewis-rag,
author = {Lewis, Patrick and Perez, Ethan and Piktus, Aleksandra and Petroni, Fabio and Karpukhin, Vladimir and Goyal, Naman and K\"{u}ttler, Heinrich and Lewis, Mike and Yih, Wen-tau and Rockt\"{a}schel, Tim and Riedel, Sebastian and Kiela, Douwe},
title = {Retrieval-augmented generation for knowledge-intensive NLP tasks},
year = {2020},
isbn = {9781713829546},
publisher = {Curran Associates Inc.},
address = {Red Hook, NY, USA},
booktitle = {Proceedings of the 34th International Conference on Neural Information Processing Systems},
articleno = {793},
numpages = {16},
location = {Vancouver, BC, Canada},
series = {NIPS '20}
}

@inproceedings{guo-etal-2025-lightrag,
    title = "{L}ight{RAG}: Simple and Fast Retrieval-Augmented Generation",
    author = "Guo, Zirui  and
      Xia, Lianghao  and
      Yu, Yanhua  and
      Ao, Tu  and
      Huang, Chao",
    editor = "Christodoulopoulos, Christos  and
      Chakraborty, Tanmoy  and
      Rose, Carolyn  and
      Peng, Violet",
    booktitle = "Findings of the Association for Computational Linguistics: EMNLP 2025",
    month = nov,
    year = "2025",
    address = "Suzhou, China",
    publisher = "Association for Computational Linguistics",
    url = "https://aclanthology.org/2025.findings-emnlp.568/",
    doi = "10.18653/v1/2025.findings-emnlp.568",
    pages = "10746--10761",
    ISBN = "979-8-89176-335-7"
}

@inproceedings{yuan-rag-selfreason,
author = {Xia, Yuan and Zhou, Jingbo and Shi, Zhenhui and Chen, Jun and Huang, Haifeng},
title = {Improving retrieval augmented language model with self-reasoning},
year = {2025},
isbn = {978-1-57735-897-8},
publisher = {AAAI Press},
url = {https://doi.org/10.1609/aaai.v39i24.34743},
doi = {10.1609/aaai.v39i24.34743},
booktitle = {Proceedings of the Thirty-Ninth AAAI Conference on Artificial Intelligence and Thirty-Seventh Conference on Innovative Applications of Artificial Intelligence and Fifteenth Symposium on Educational Advances in Artificial Intelligence},
articleno = {2845},
numpages = {9},
series = {AAAI'25/IAAI'25/EAAI'25}
}

@article{huang-rag,
author = {Huang, Yizheng and Huang, Jimmy Xiangji},
title = {A Survey on Retrieval-Augmented Text Generation for Large Language Models},
year = {2026},
issue_date = {September 2026},
publisher = {Association for Computing Machinery},
address = {New York, NY, USA},
volume = {58},
number = {12},
issn = {0360-0300},
url = {https://doi.org/10.1145/3805774},
doi = {10.1145/3805774},
journal = {ACM Comput. Surv.},
month = may,
articleno = {300},
numpages = {38}
}

@misc{liu2025comprehensivesurveylongcontext,
      title={A Comprehensive Survey on Long Context Language Modeling}, 
      author={Jiaheng Liu and Dawei Zhu and Zhiqi Bai and Yancheng He and Huanxuan Liao and Haoran Que and Zekun Wang and Chenchen Zhang and Ge Zhang and Jiebin Zhang and Yuanxing Zhang and Zhuo Chen and Hangyu Guo and Shilong Li and Ziqiang Liu and Yong Shan and Yifan Song and Jiayi Tian and Wenhao Wu and Zhejian Zhou and Ruijie Zhu and Junlan Feng and Yang Gao and Shizhu He and Zhoujun Li and Tianyu Liu and Fanyu Meng and Wenbo Su and Yingshui Tan and Zili Wang and Jian Yang and Wei Ye and Bo Zheng and Wangchunshu Zhou and Wenhao Huang and Sujian Li and Zhaoxiang Zhang},
      year={2025},
      eprint={2503.17407},
      archivePrefix={arXiv},
      primaryClass={cs.CL},
      url={https://arxiv.org/abs/2503.17407}, 
}

@inproceedings{
hu2022lora,
title={Lo{RA}: Low-Rank Adaptation of Large Language Models},
author={Edward J Hu and yelong shen and Phillip Wallis and Zeyuan Allen-Zhu and Yuanzhi Li and Shean Wang and Lu Wang and Weizhu Chen},
booktitle={International Conference on Learning Representations},
year={2022},
url={https://openreview.net/forum?id=nZeVKeeFYf9}
}

@misc{
shuttleworth2025lora,
title={Lo{RA} vs Full Fine-tuning: An Illusion of Equivalence},
author={Reece S Shuttleworth and Jacob Andreas and Antonio Torralba and Pratyusha Sharma},
year={2025},
url={https://openreview.net/forum?id=PGNdDfsI6C}
}

@inproceedings{lester-etal-2021-prompttuning,
    title = "The Power of Scale for Parameter-Efficient Prompt Tuning",
    author = "Lester, Brian  and
      Al-Rfou, Rami  and
      Constant, Noah",
    editor = "Moens, Marie-Francine  and
      Huang, Xuanjing  and
      Specia, Lucia  and
      Yih, Scott Wen-tau",
    booktitle = "Proceedings of the 2021 Conference on Empirical Methods in Natural Language Processing",
    month = nov,
    year = "2021",
    address = "Online and Punta Cana, Dominican Republic",
    publisher = "Association for Computational Linguistics",
    url = "https://aclanthology.org/2021.emnlp-main.243/",
    doi = "10.18653/v1/2021.emnlp-main.243",
    pages = "3045--3059"
}

@misc{jiang2025knowmerespondme,
      title={Know Me, Respond to Me: Benchmarking LLMs for Dynamic User Profiling and Personalized Responses at Scale}, 
      author={Bowen Jiang and Zhuoqun Hao and Young-Min Cho and Bryan Li and Yuan Yuan and Sihao Chen and Lyle Ungar and Camillo J. Taylor and Dan Roth},
      year={2025},
      eprint={2504.14225},
      archivePrefix={arXiv},
      primaryClass={cs.CL},
      url={https://arxiv.org/abs/2504.14225}, 
}

@inproceedings{maharana-etal-2024-evaluating,
    title = "Evaluating Very Long-Term Conversational Memory of {LLM} Agents",
    author = "Maharana, Adyasha  and
      Lee, Dong-Ho  and
      Tulyakov, Sergey  and
      Bansal, Mohit  and
      Barbieri, Francesco  and
      Fang, Yuwei",
    editor = "Ku, Lun-Wei  and
      Martins, Andre  and
      Srikumar, Vivek",
    booktitle = "Proceedings of the 62nd Annual Meeting of the Association for Computational Linguistics (Volume 1: Long Papers)",
    month = aug,
    year = "2024",
    address = "Bangkok, Thailand",
    publisher = "Association for Computational Linguistics",
    url = "https://aclanthology.org/2024.acl-long.747/",
    doi = "10.18653/v1/2024.acl-long.747",
    pages = "13851--13870"
}

@misc{wang2025mextendingmemoryllmscalable,
      title={M+: Extending MemoryLLM with Scalable Long-Term Memory}, 
      author={Yu Wang and Dmitry Krotov and Yuanzhe Hu and Yifan Gao and Wangchunshu Zhou and Julian McAuley and Dan Gutfreund and Rogerio Feris and Zexue He},
      year={2025},
      eprint={2502.00592},
      archivePrefix={arXiv},
      primaryClass={cs.CL},
      url={https://arxiv.org/abs/2502.00592}, 
}

@misc{liu2022fewshotparameterefficientfinetuningbetter,
      title={Few-Shot Parameter-Efficient Fine-Tuning is Better and Cheaper than In-Context Learning}, 
      author={Haokun Liu and Derek Tam and Mohammed Muqeeth and Jay Mohta and Tenghao Huang and Mohit Bansal and Colin Raffel},
      year={2022},
      eprint={2205.05638},
      archivePrefix={arXiv},
      primaryClass={cs.LG},
      url={https://arxiv.org/abs/2205.05638}, 
}

@inproceedings{
fang2026lightmem,
title={LightMem: Lightweight and Efficient Memory-Augmented Generation},
author={Jizhan Fang and Xinle Deng and Haoming Xu and Ziyan Jiang and Yuqi Tang and Ziwen Xu and Shumin Deng and Yunzhi Yao and Mengru Wang and Shuofei Qiao and Huajun Chen and Ningyu Zhang},
booktitle={The Fourteenth International Conference on Learning Representations},
year={2026},
url={https://openreview.net/forum?id=dyJ0GWpjJB}
}

@misc{huber2025embeddingtoprefixparameterefficientpersonalizationpretrained,
      title={Embedding-to-Prefix: Parameter-Efficient Personalization for Pre-Trained Large Language Models}, 
      author={Bernd Huber and Ghazal Fazelnia and Andreas Damianou and Sebastian Peleato and Max Lefarov and Praveen Ravichandran and Marco De Nadai and Mounia Lalmas-Roellke and Paul N. Bennett},
      year={2025},
      eprint={2505.17051},
      archivePrefix={arXiv},
      primaryClass={cs.CL},
      url={https://arxiv.org/abs/2505.17051}, 
}

@misc{mcinnes2020umapuniformmanifoldapproximation,
      title={UMAP: Uniform Manifold Approximation and Projection for Dimension Reduction}, 
      author={Leland McInnes and John Healy and James Melville},
      year={2020},
      eprint={1802.03426},
      archivePrefix={arXiv},
      primaryClass={stat.ML},
      url={https://arxiv.org/abs/1802.03426}, 
}

@misc{tandon2025endtoendtesttimetraininglong,
      title={End-to-End Test-Time Training for Long Context}, 
      author={Arnuv Tandon and Karan Dalal and Xinhao Li and Daniel Koceja and Marcel Rød and Sam Buchanan and Xiaolong Wang and Jure Leskovec and Sanmi Koyejo and Tatsunori Hashimoto and Carlos Guestrin and Jed McCaleb and Yejin Choi and Yu Sun},
      year={2025},
      eprint={2512.23675},
      archivePrefix={arXiv},
      primaryClass={cs.LG},
      url={https://arxiv.org/abs/2512.23675}, 
}

@misc{chhikara2025mem0buildingproductionreadyai,
      title={Mem0: Building Production-Ready AI Agents with Scalable Long-Term Memory}, 
      author={Prateek Chhikara and Dev Khant and Saket Aryan and Taranjeet Singh and Deshraj Yadav},
      year={2025},
      eprint={2504.19413},
      archivePrefix={arXiv},
      primaryClass={cs.CL},
      url={https://arxiv.org/abs/2504.19413}, 
}

@misc{wang2024memoryllmselfupdatablelargelanguage,
      title={MEMORYLLM: Towards Self-Updatable Large Language Models}, 
      author={Yu Wang and Yifan Gao and Xiusi Chen and Haoming Jiang and Shiyang Li and Jingfeng Yang and Qingyu Yin and Zheng Li and Xian Li and Bing Yin and Jingbo Shang and Julian McAuley},
      year={2024},
      eprint={2402.04624},
      archivePrefix={arXiv},
      primaryClass={cs.CL},
      url={https://arxiv.org/abs/2402.04624}, 
}

@inproceedings{
zhang2026memgen,
title={MemGen: Weaving Generative Latent Memory for Self-Evolving Agents},
author={Guibin Zhang and Muxin Fu and Shuicheng YAN},
booktitle={The Fourteenth International Conference on Learning Representations},
year={2026},
url={https://openreview.net/forum?id=vI56m4Iu4e}
}

@article{memory3-yang,
   title={Memory$^3$: Language Modeling with Explicit Memory},
   volume={3},
   ISSN={2790-203X},
   url={http://dx.doi.org/10.4208/jml.240708},
   DOI={10.4208/jml.240708},
   number={3},
   journal={Journal of Machine Learning},
   publisher={Global Science Press},
   author={Yang, Hongkang and Lin, Zehao and Wang, Wenjin and Wu, Hao and Li, Zhiyu and Tang, Bo and Wei, Wenqiang and Wang, Jinbo and Tang, Zeyun and Song, Shichao and Xi, Chenyang and Yu, Yu and Chen, Kai and Xiong, Feiyu and Tang, Linpeng and E, Weinan},
   year={2024},
   month=Sept, pages={300–346} }

@misc{
wang2026memalpha,
title={{MEM}-\${\textbackslash}alpha\$: {LEARNING} {MEMORY} {CONSTRUCTION} {VIA} {REINFORCEMENT} {LEARNING}},
author={Yu Wang and Ryuichi Takanobu and Zhiqi Liang and Yuzhen Mao and Yuanzhe Hu and Julian McAuley and Xiaojian Wu},
year={2026},
url={https://openreview.net/forum?id=dm42omwep1}
}

@misc{wang2025mirixmultiagentmemoryllmbased,
      title={MIRIX: Multi-Agent Memory System for LLM-Based Agents}, 
      author={Yu Wang and Xi Chen},
      year={2025},
      eprint={2507.07957},
      archivePrefix={arXiv},
      primaryClass={cs.CL},
      url={https://arxiv.org/abs/2507.07957}, 
}

@misc{yan2026memoryr1enhancinglargelanguage,
      title={Memory-R1: Enhancing Large Language Model Agents to Manage and Utilize Memories via Reinforcement Learning}, 
      author={Sikuan Yan and Xiufeng Yang and Zuchao Huang and Ercong Nie and Zifeng Ding and Zonggen Li and Xiaowen Ma and Jinhe Bi and Kristian Kersting and Jeff Z. Pan and Hinrich Schütze and Volker Tresp and Yunpu Ma},
      year={2026},
      eprint={2508.19828},
      archivePrefix={arXiv},
      primaryClass={cs.CL},
      url={https://arxiv.org/abs/2508.19828}, 
}

@misc{hu2026memoryageaiagents,
      title={Memory in the Age of AI Agents}, 
      author={Yuyang Hu and Shichun Liu and Yanwei Yue and Guibin Zhang and Boyang Liu and Fangyi Zhu and Jiahang Lin and Honglin Guo and Shihan Dou and Zhiheng Xi and Senjie Jin and Jiejun Tan and Yanbin Yin and Jiongnan Liu and Zeyu Zhang and Zhongxiang Sun and Yutao Zhu and Hao Sun and Boci Peng and Zhenrong Cheng and Xuanbo Fan and Jiaxin Guo and Xinlei Yu and Zhenhong Zhou and Zewen Hu and Jiahao Huo and Junhao Wang and Yuwei Niu and Yu Wang and Zhenfei Yin and Xiaobin Hu and Yue Liao and Qiankun Li and Kun Wang and Wangchunshu Zhou and Yixin Liu and Dawei Cheng and Qi Zhang and Tao Gui and Shirui Pan and Yan Zhang and Philip Torr and Zhicheng Dou and Ji-Rong Wen and Xuanjing Huang and Yu-Gang Jiang and Shuicheng Yan},
      year={2026},
      eprint={2512.13564},
      archivePrefix={arXiv},
      primaryClass={cs.CL},
      url={https://arxiv.org/abs/2512.13564}, 
}

@inproceedings{tan-etal-2024-lloco,
    title = "{LL}o{CO}: Learning Long Contexts Offline",
    author = "Tan, Sijun  and
      Li, Xiuyu  and
      Patil, Shishir G  and
      Wu, Ziyang  and
      Zhang, Tianjun  and
      Keutzer, Kurt  and
      Gonzalez, Joseph E.  and
      Popa, Raluca Ada",
    editor = "Al-Onaizan, Yaser  and
      Bansal, Mohit  and
      Chen, Yun-Nung",
    booktitle = "Proceedings of the 2024 Conference on Empirical Methods in Natural Language Processing",
    month = nov,
    year = "2024",
    address = "Miami, Florida, USA",
    publisher = "Association for Computational Linguistics",
    url = "https://aclanthology.org/2024.emnlp-main.975/",
    doi = "10.18653/v1/2024.emnlp-main.975",
    pages = "17605--17621"
}

@article{Snell2022Learningcontext,
  title={Learning by Distilling Context},
  author={Charles Burton Snell and Dan Klein and Ruiqi Zhong},
  journal={ArXiv},
  year={2022},
  volume={abs/2209.15189},
  url={https://api.semanticscholar.org/CorpusID:252668389}
}

@misc{zoph2022stmoedesigningstabletransferable,
      title={ST-MoE: Designing Stable and Transferable Sparse Expert Models}, 
      author={Barret Zoph and Irwan Bello and Sameer Kumar and Nan Du and Yanping Huang and Jeff Dean and Noam Shazeer and William Fedus},
      year={2022},
      eprint={2202.08906},
      archivePrefix={arXiv},
      primaryClass={cs.CL},
      url={https://arxiv.org/abs/2202.08906}, 
}

@misc{li2025dynmoleboostingmixturelora,
      title={DynMoLE: Boosting Mixture of LoRA Experts Fine-Tuning with a Hybrid Routing Mechanism}, 
      author={Dengchun Li and Naizheng Wang and Zihao Zhang and Haoyang Yin and Lei Duan and Meng Xiao and Mingjie Tang},
      year={2025},
      eprint={2504.00661},
      archivePrefix={arXiv},
      primaryClass={cs.CL},
      url={https://arxiv.org/abs/2504.00661}, 
}

\appendix

\newpage
\clearpage

\section{Appendix}

\subsection{Dataset full details}
\label{app:datasets}
\begin{table}[h]
\centering
\small
\setlength{\tabcolsep}{6pt}
\resizebox{0.3\textwidth}{!}{%
\begin{tabular}{llc}
\toprule
ID & Category & \# Questions \\
\midrule
1 & Multi-hop    & 282 \\
2 & Temporal     & 321 \\
3 & Open-domain  & 96  \\
4 & Single-hop   & 841 \\
\midrule
  & \textbf{Total} & \textbf{1540} \\
\bottomrule
\end{tabular}%
}
\caption{LoCOMO per-category question counts. 1. Multi-hop questions
require synthesizing information from multiple sessions. 2. temporal
questions require time-related reasoning. 3. open-domain questions
require commonsense or external knowledge. 4. single-hop questions are
answerable from a single session.}
\label{tab:locomo-stats}
\end{table}

\begin{table}[h]
\centering
\small
\setlength{\tabcolsep}{6pt}
\resizebox{0.5\textwidth}{!}{%
\begin{tabular}{lc}
\toprule
Category & \# Questions \\
\midrule
Recall user shared facts \textbf{(R-Fact)} & 146 \\
Suggest new ideas \textbf{(Sug)}                                & 93  \\
Track full preference evolution \textbf{(P-Evol)}                   & 139 \\
Recall reasons behind previous updates \textbf{(P-Rsn)}           & 99  \\
Provide preference-aligned recommendations \textbf{(P-Rec)}       & 55  \\
Generalize to new scenarios \textbf{(Gen)}                      & 57  \\
\midrule
\textbf{Total}                                    & \textbf{589} \\
\bottomrule
\end{tabular}%
}
\caption{PersonaMem v1 per-category question counts.}
\label{tab:personamem-stats}
\end{table}

% =====================================================================
% APPENDIX: CHAT TEMPLATE AND TRAINING DETAILS
% =====================================================================
\subsection{Chat Template and Training Details}
\label{app:chat-template}

We train \ourmethod{} with a chunk-based objective that operates over a
chat-formatted sequence. The soft prompts produced by the memory module
are inserted into the first system block, immediately after the system
token and before the
"\texttt{\#\#\# CONTEXT END \#\#\#}" marker that delimits the latent
context from the textual instruction. The training loss is computed
only from the first context token. All tokens
preceding it (including the soft-prompt positions, the system
instruction, and the user instruction) are masked out and do not
contribute to the loss. This ensures that the model is optimized to
\emph{use the latent memory} rather than to reconstruct the template
scaffolding around it. The exact chat template used during training is reproduced below.

\begin{table}[h]
\centering
\small
\renewcommand{\arraystretch}{1.3}
\begin{tabular}{@{}p{0.13\linewidth}p{0.80\linewidth}@{}}
\toprule
\textbf{Role} & \textbf{Content} \\
\midrule
\texttt{system} &
\textsl{[Dynamic soft prompts from all heads]}\;\texttt{\#\#\# CONTEXT END \#\#\#}\par
You have access to a latent memory block above. This block contains
relevant context, the end of which is marked with
\texttt{\#\#\# CONTEXT END \#\#\#}. \\
\midrule
\texttt{user} &
Prioritize the information in the above latent memory block to
predict the following next text accurately. \\
\midrule
\texttt{assistant} &
\textsl{[Context chunk tokens - loss is computed here]} \\
\bottomrule
\end{tabular}
\caption{Message structure of the chat template used for chunk-based
training. Soft prompts are prepended inside the system block,
immediately before the \texttt{\#\#\# CONTEXT END \#\#\#} marker. The
training loss is masked everywhere except on the assistant turn,
starting from the first context token.}
\label{tab:chat-template}
\end{table}

Two aspects of this template are worth highlighting. Firstly, the soft
prompts are placed inside the \emph{system block} rather than as a
separate prefix, which keeps them within the region of the sequence that
the underlying chat model is trained to treat as conditioning. Secondly,
the user instruction is kept short and generic to anchor the assistant block but not encode any task-specific information. 

% =====================================================================
% APPENDIX: LOCOMO EVALUATION JUDGE PROMPT
% =====================================================================
\subsection{LOCOMO Evaluation Judge Prompt}
\label{app:locomo-judge}

For evaluation on LOCOMO we follow the protocol introduced Mem0 \cite{chhikara2025mem0buildingproductionreadyai} and use an LLM-as-a-judge setup. Each generated answer is compared
against the gold answer by an GPT-5-mini that returns a
binary CORRECT / WRONG label together with a short rationale. The judge
is prompted to be lenient with respect to surface form (e.g.\ verbosity,
paraphrase, alternative date formats) and to treat refusals as correct
when the gold answer itself indicates that the information is not
available in the conversation history. We provide both the system
prompt and the per-example user prompt below for reproducibility.

\begin{promptbox}[title=System prompt]
\small
You are an LLM Judge. Your task is to label an answer to a question as
\texttt{CORRECT} or \texttt{WRONG}. You will be given the following
data:
\begin{enumerate}[topsep=2pt,itemsep=0pt,leftmargin=*]
  \item a question (posed by one user to another user),
  \item a ``gold'' (ground truth) answer,
  \item a generated answer,
\end{enumerate}
which you will score as \texttt{CORRECT}/\texttt{WRONG}.
 
\medskip
The point of the question is to ask about something one user should
know about the other user based on their prior conversations.
 
\medskip
The gold answer will usually be a concise and short answer that
includes the referenced topic, for example:
\begin{quote}
\textbf{Question:} Do you remember what I got the last time I went to
Hawaii?\\
\textbf{Gold answer:} A shell necklace
\end{quote}
The generated answer might be much longer, but you should be generous
with your grading---as long as it touches on the same topic as the
gold answer, it should be counted as \texttt{CORRECT}.
 
\medskip
For time-related questions, the gold answer will be a specific date,
month, year, etc. The generated answer might be much longer or use
relative time references (like ``last Tuesday'' or ``next month''),
but you should be generous with your grading---as long as it refers
to the same date or time period as the gold answer, it should be
counted as \texttt{CORRECT}. Even if the format differs (e.g.,
``May 7th'' vs.\ ``7 May''), consider it \texttt{CORRECT} if it is
the same date.
 
\medskip
There is an edge case where the actual answer cannot be found in the
data, and in that case the gold answer will say so (e.g.,\ ``You did
not mention this information.''); if the generated answer says that
it cannot be answered or it does not know all the details, it should
be counted as \texttt{CORRECT}.
 
\medskip
Respond with JSON:
\texttt{\{"reasoning": "...", "correct": true or false\}}
\end{promptbox}

\begin{promptbox}[title=User prompt (per example)]
\small
Now it's time for the real question:
\begin{quote}
\textbf{Question:} \texttt{\{question\}}\\
\textbf{Gold answer:} \texttt{\{answer\}}\\
\textbf{Generated answer:} \texttt{\{model\_output\}}
\end{quote}
First, provide a short (one sentence) explanation of your reasoning.
Short reasoning is preferred. If it's correct, set
\texttt{correct=true}.
 
\medskip
Respond with JSON:
\texttt{\{"reasoning": "...", "correct": true or false\}}
\end{promptbox}

The judge is queried independently for every (question, gold answer,
generated answer) triple, and final accuracies are reported as the
fraction of triples labelled CORRECT. We use the same judge model and
decoding settings across all methods compared in the main paper to
ensure a like-for-like comparison.

\section{Rubric-Based User Similarity Evaluation}
\label{app:user_similarity}

To interpret the structure observed in the learned memory slot embeddings (Figure~\ref{fig:umap_slots}), we conduct a qualitative similarity analysis across representative users from PersonaMemV1. The objective of this study is to determine whether the learned latent representations preserve meaningful behavioral distinctions between users beyond topical overlap.

\subsection{Evaluation Setup}
\label{sec:evalrubric}
For each evaluated user, the full PersonaMemV1 conversational history ($\sim$32K tokens) was provided to \textsc{Claude Opus 4.7} as a judge model. The judge was instructed to compare users pairwise using a structured similarity rubric designed to capture long-term behavioral and interactional characteristics. The details of the rubric is provided in Table \ref{tab:rubric_dimensions}.
\begin{table*}[t]
\centering
\small
\setlength{\tabcolsep}{5pt}
\renewcommand{\arraystretch}{1.12}
\begin{tabular}{p{0.5cm}p{3.6cm}p{9.2cm}}
\toprule
\textbf{\#} & \textbf{Dimension} & \textbf{Evaluation Criteria} \\
\midrule

1 &
Domain-specific likes &
Measures overlap in what users actively enjoy or seek out, including genres, hobbies, activities, products, aesthetics, media preferences, and recurring interests. Higher scores require overlap at a granular level rather than broad category similarity alone. \\

2 &
Domain-specific dislikes &
Measures overlap in what users avoid, reject, or dislike. Shared dislikes are treated as strong behavioral signals and often correlate more strongly with personality alignment than shared positive preferences. \\

3 &
Lifestyle and routines &
Evaluates similarity in routines and behavioral structure, including work style, social rhythms, travel habits, daily routines, activity patterns, and preference for structured versus casual engagement. \\

4 &
Values and priorities &
Measures alignment in what users consistently prioritize, such as learning, career growth, authenticity, financial security, community engagement, autonomy, or personal relationships. \\

5 &
Personality and communication style &
Evaluates conversational tone and behavioral traits, including verbosity, emotional expressiveness, openness, humor, decisiveness, analytical thinking, and interaction style. \\

6 &
Goals and aspirations &
Measures similarity in long-term ambitions and pursuits, including professional goals, learning objectives, relationship goals, creative aspirations, or community-building efforts. \\

7 &
Constraints and circumstances &
Captures shared constraints or contextual factors such as occupational limitations, financial constraints, family situations, health considerations, or geographic circumstances. \\

8 &
Expertise and knowledge domains &
Measures overlap in areas of expertise, technical knowledge, cultural fluency, or domains where users demonstrate substantial familiarity or long-term engagement. \\

9 &
Aesthetic and stylistic preferences &
Evaluates alignment in stylistic preferences, including visual aesthetics, design sensibilities, narrative preferences, fashion tastes, media consumption style, and environmental preferences. \\

10 &
Preference evolution and behavioral stability &
Measures whether users exhibit similar behavioral trajectories over time, such as recurring exploration patterns, preference shifts, adoption or abandonment of activities, and long-term interaction dynamics. \\

\bottomrule
\end{tabular}
\caption{Dimensions used in the rubric-based similarity evaluation. The rubric is designed to capture long-term behavioral and interactional similarity beyond surface-level topical overlap.}
\label{tab:rubric_dimensions}
\end{table*}

\begin{table}[t]
\centering
\small
\setlength{\tabcolsep}{8pt}
\renewcommand{\arraystretch}{1.12}
\resizebox{\linewidth}{!}{%
\begin{tabular}{p{3.8cm}p{3.5cm}}
\toprule
\textbf{Mean Similarity Score} & \textbf{Qualitative Label} \\
\midrule
3.5 -- 4.0 & Highly similar \\
2.5 -- 3.4 & Moderately similar \\
1.5 -- 2.4 & Somewhat similar \\
0.5 -- 1.4 & Mostly dissimilar \\
0.0 -- 0.4 & Dissimilar / opposed \\
\bottomrule
\end{tabular}
}
\caption{Qualitative interpretation of aggregate similarity scores. The overall similarity score is computed as the mean across all dimensions.}
\label{tab:overall_similarity_labels}
\end{table}

\noindent\textbf{Aggregate scoring.}
The overall similarity score is computed as the mean across all dimensions. The interpretation of the aggregate similarity scores is provided in Table \ref{tab:overall_similarity_labels}.

\noindent\textbf{Evaluation guidelines.}
The rubric prioritizes evidence density and specificity during scoring. Explicitly stated preferences are weighted more strongly than inferred preferences, and granular overlaps (e.g., highly specific shared behaviors or preferences) are scored higher than broad category-level similarities. Partial overlaps are assigned intermediate scores when one user's preferences form only a subset of another user's broader behavioral patterns.

\subsection{Representative Pairwise Similarity Scores}

\subsection{Qualitative Findings}
\label{qual-findings}
A consistent pattern emerges from the analysis. Users~2, 4, and 15 form a relatively coherent behavioral cluster despite discussing substantially different topical domains. The judge model identifies several recurring similarities across these users:
\begin{itemize}[leftmargin=15pt]
    \item personality and communication styles,
    \item repeated engagement with organized communities or activities,
    \item preference shifts toward smaller and more authentic interactions over time,
    \item recurring cycles of exploration, burnout, and re-engagement.
\end{itemize}

Importantly, these similarities arise primarily from interaction dynamics and behavioral trajectories rather than explicit topical overlap. This suggests that the learned memory slots capture higher-level conversational structure beyond surface-level semantic similarity. An example output from the judge model is presented in Table \ref{tab:most_similar_pair}. 

In contrast, User~14 is consistently identified as the most behaviorally distinct user. The output from the judge model on the most dissimilar user pairs (14 and 15) is presented in Table \ref{tab:most_dissimilar_pair}. All pairings involving User~14 receive the lowest similarity scores in the evaluation. The judge model characterizes User~14 as more socially exploratory and emotionally driven compared to the remaining users, whose interactions are primarily intellectual, educational, or community-oriented. This distinction is also reflected in the learned embedding space visualized in Figure~\ref{fig:umap_slots}, where User~14 occupies a comparatively isolated region. The agreement between the qualitative rubric analysis and the geometric structure of the learned memory slots provides evidence that the model preserves meaningful long-term behavioral distinctions between users.

\begin{table*}[t]
\centering
\small
\setlength{\tabcolsep}{5pt}
\renewcommand{\arraystretch}{1.12}
\begin{tabular}{p{3.0cm}p{0.8cm}p{9.6cm}}
\toprule
\textbf{Dimension} & \textbf{Score} & \textbf{Judge Model Output (User 2 $\leftrightarrow$ User 4)} \\
\midrule

Domain-specific likes &
2 &
No topical overlap (books vs. films), but both users strongly engage with narrative-driven media, value indie creators, and participate in cultural events and communities. \\

Domain-specific dislikes &
4 &
Strong alignment in disliking rigid club formats, overwhelming large-scale events, binge-style consumption, and shallow online discussions. \\

Lifestyle and routines &
3 &
Both users shifted from structured participation toward more casual and self-paced interactions over time. \\

Values and priorities &
3 &
Both value authentic engagement, learning, community participation, and depth of experience over performative participation. \\

Personality and communication style &
4 &
Highly reflective, introspective, and articulate communication styles with strong preference for intimate interactions over crowded environments. \\

Goals and aspirations &
3 &
Both aim to deepen their craft and engage more authentically with their respective communities. \\

Expertise and knowledge domains &
2 &
Different topical domains, but both exhibit deep cultural fluency and creator-oriented engagement patterns. \\

Aesthetic and stylistic preferences &
3 &
Shared preference for emotional depth, character-driven narratives, and non-mainstream experiences. \\

Preference evolution and stability &
4 &
Nearly identical behavioral trajectories characterized by exploration, burnout from structured engagement, and eventual transition toward smaller authentic interactions. \\

\midrule
\textbf{Overall} &
\textbf{3.11} &
\textit{``Despite zero topical overlap, this pair is the most behaviorally aligned. They share remarkably similar interaction trajectories and conversational styles.''} \\

\bottomrule
\end{tabular}
\caption{Judge model outputs using Claude 4.7-Opus for the most similar user pair User 2 and 4 identified during rubric-based evaluation. Despite discussing different domains, the users exhibit highly aligned behavioral patterns, communication styles, and interaction trajectories.}
\label{tab:most_similar_pair}
\end{table*}

\begin{table*}[t]
\centering
\small
\setlength{\tabcolsep}{5pt}
\renewcommand{\arraystretch}{1.12}
\begin{tabular}{p{3.0cm}p{0.8cm}p{9.6cm}}
\toprule
\textbf{Dimension} & \textbf{Score} & \textbf{Judge Model Output (User 15 $\leftrightarrow$ User 14)} \\
\midrule

Domain-specific likes &
1 &
Very different interests and motivations. Slight overlap only in attending community discussions or social events. \\

Domain-specific dislikes &
2 &
Limited overlap. Both dislike unproductive networking events, but for substantially different reasons: inefficacy versus social discomfort. \\

Lifestyle and routines &
2 &
Both explore many activities, but their motivations differ substantially (social connection vs. knowledge-building). \\

Values and priorities &
2 &
Only partial overlap. One prioritizes personal connection while the other prioritizes education and financial empowerment. \\

Personality and communication style &
3 &
Both are reflective and verbose, but one is emotionally exploratory while the other is analytically self-evaluative. \\

Goals and aspirations &
1 &
Fundamentally different long-term goals and motivations. \\

Expertise and knowledge domains &
1 &
No meaningful overlap in expertise or domain knowledge. \\

Aesthetic and stylistic preferences &
2 &
Some overlap in preferring smaller and casual settings over formal or highly structured environments. \\

Preference evolution and stability &
3 &
Surface-level similarity in exploration patterns, but driven by substantially different underlying motivations. \\

\midrule
\textbf{Overall} &
\textbf{1.89} &
\textit{``The similarity is largely superficial and behavioral rather than personality-level alignment.''} \\

\bottomrule
\end{tabular}
\caption{Judge model outputs using Claude 4.7-Opus for the most dissimilar user pair User 14 and 15, identified during rubric-based evaluation. Although some high-level behavioral patterns overlap, the users differ substantially in motivations, conversational dynamics, and long-term goals.}
\label{tab:most_dissimilar_pair}
\end{table*}

\section{Qualitative Examples on LoCOMO Dataset}
\label{app:qualitative}

Table~\ref{tab:locomo_success} shows representative successes and
Table~\ref{tab:locomo_failure} shows characteristic failure modes.

\subsubsection{Success Cases}
The model demonstrates strong recall of semantic and relational content across a range of question types. On multi-hop questions, it correctly chains facts distributed across sessions. On single-hop questions, there are examples where important entities are reliably recovered even from long contexts. Notably, the model also \textit{generalizes beyond direct recall, inferring plausible continuations from memorized personality traits} and values, suggesting that the latent memory \textit{encodes not just surface facts but a richer implicit representation of the user's profile}.

\begin{table*}[h!]
\centering
\small
\caption{LoCOMO success cases. The model answers correctly from latent
memory alone, with no conversation context at inference time.}
\label{tab:locomo_success}
\renewcommand{\arraystretch}{1.4}
\setlength{\tabcolsep}{5pt}
\resizebox{0.8\textwidth}{!}{
\begin{tabular}{>{\raggedright}p{2.2cm}
                >{\raggedright}p{3.8cm}
                >{\raggedright}p{2.8cm}
                >{\raggedright\arraybackslash}p{4.5cm}}
\toprule
\textbf{Type} & \textbf{Question} & \textbf{Gold Answer} & \textbf{Model Response} \\
\midrule

Multi-hop &
``What places has Joanna submitted her work to?'' &
film contest, film festival &
``Joanna has submitted her work to a few different places, including a few
film festivals and some online platforms.''

\smallskip
{\small\textit{Captures the key entity (film festivals) and plausibly
generalizes to online platforms. It is successful in chaining facts
across sessions.}} \\

\midrule

Single-hop \newline (partial recall) &
``What authors has Tim read books from?'' &
J.K.\ Rowling, G.R.R.\ Martin, P.\ Rothfuss, P.\ Coelho, J.R.R.\ Tolkien &
``Tim has read books from J.K.\ Rowling, George R.R.\ Martin.''

\smallskip
{\small\textit{A correct but compressed subset: high-salience entities are
preserved while less-emphasised ones are dropped.}} \\

\midrule

Open-domain/Reasoning &
``What fields would Caroline be likely to pursue in her education?'' &
psychology, counseling certification &
``Caroline would likely pursue education in psychology, counseling, or social
work. She's passionate about advocating for the LGBTQ+ community and helping
others.''

\smallskip
{\small\textit{The model does not merely recall a stated fact. It infers a
plausible educational trajectory from memorized personality traits and
values.}} \\

\midrule

Factual recall &
``Where did Caroline move from 4 years ago?'' &
Sweden &
``Caroline moved from Sweden to Canada 4 years ago.''

\smallskip
{\small\textit{Precise single-entity recall of a static biographical detail
buried across multiple sessions.}} \\

\bottomrule
\end{tabular}
}
\end{table*}

\subsubsection{Failure Cases}

The compressed latent representation is lossy for \emph{precise numerical
and temporal content such as exact dates, counts, and verbatim strings}, while the semantic content is largely preserved.

\begin{table*}[t!]
\centering
\small
\caption{LoCOMO failure cases. The dominant failure mode is imprecise
numerical or temporal content despite correct semantic recall.}
\label{tab:locomo_failure}
\renewcommand{\arraystretch}{1.4}
\setlength{\tabcolsep}{5pt}
\resizebox{0.7\textwidth}{!}{
\begin{tabular}{>{\raggedright}p{2.2cm}
                >{\raggedright}p{3.8cm}
                >{\raggedright}p{2.0cm}
                >{\raggedright\arraybackslash}p{5.3cm}}
\toprule
\textbf{Type} & \textbf{Question} & \textbf{Gold Answer} & \textbf{Model Response} \\
\midrule

Exact-date recall &
``When did Caroline go to the LGBTQ support group?'' &
7 May 2023 &
``Caroline went to the LGBTQ support group in June 2023.''

\smallskip
{\small\textit{Correct semantic content (approximate month and year),
imprecise numerical content (wrong month, no day). Timestamps are the
first detail compressed away.}} \\

\midrule

Exact-count recall &
``How many Prius has Evan owned?'' &
two &
``Evan had one Prius. It's been a great way to reduce his environmental
impact.''

\smallskip
{\small\textit{The model retains the entity (Prius) and the associated
sentiment (environmental motivation) but loses the exact count. Small
integers embedded in incidental context are prone to compression error.}} \\

\bottomrule
\end{tabular}
}
\end{table*}

These results suggest a clear direction for future work by augmenting
the latent memory with a lightweight cache or a scratchpad for high-precision numerical and temporal facts, while retaining the
compact semantic compression for broader contextual knowledge. In this way, it can potentially help get the best of both worlds (RAG \& latent memory).

\section{Personamemv1 Details}
\subsection{Prompt used for evaluation}

The prompt used for Personamem evaluation is presented in Figure \ref{pmem-prompt}.
\begin{figure}[h!]
\caption{PersonaMemv1 inference prompt. The three fields are concatenated
(separated by blank lines) and wrapped in a Qwen3 user-turn chat template.
No long-form conversation context or user history is prepended.}
\lstset{
  basicstyle=\small\ttfamily,
  breaklines=true,
  frame=single,
  backgroundcolor=\color{gray!10},
  xleftmargin=6pt, xrightmargin=6pt
}
\begin{lstlisting}
Find the most appropriate model response from the options.
Pick a single option after the special token <final_answer>.
Provide the reasoning for your choice after final answer.

{user_question_or_message}

{all_options}
\end{lstlisting}
\label{pmem-prompt}
\end{figure}

\subsection{Qualitative Examples on PersonaMemv1}

We present qualitative examples from the PersonaMem 32k evaluation of our
model LPM. The examples illustrate that the model learns to compress
personalized user memory \emph{directly into its parameters}. At inference
time it \textit{successfully tracks preference evolution}, \textit{generalizes learned knowledge across domains}, and \textit{generates preference-aligned recommendations}, without having access
to the original conversation. Table \ref{tab:personamem_examples} presents qualitative examples of LPM's output on Personamemv1. This highlights the effectiveness of latent memory approaches in memorizing personalized user preferences, while being considerabily more efficient compared to full context and text memory methods.

\begin{table*}[h!]
\centering
\small
\caption{Qualitative PersonaMem 32k examples. \textbf{Correct} model outputs
are shown for three task types; a \textbf{failure} case for
\textit{Suggest New Ideas} is included in the final row.
The correct option is shown in \textbf{bold} in each option list.}
\label{tab:personamem_examples}
\renewcommand{\arraystretch}{1.4}
\setlength{\tabcolsep}{5pt}
\begin{tabular}{>{\raggedright}p{2.4cm}
                >{\raggedright}p{5.3cm}
                >{\raggedright}p{4.1cm}
                c}
\toprule
\textbf{Task Type} &
\textbf{User Query \& Options (abbreviated)} &
\textbf{Model Response (excerpt)} &
\textbf{} \\
\midrule

\textit{Track Full Preference Evolution} &
\textit{Query:} ``It delves into the psychology behind book characters and
narratives, which has really sparked my interest again\ldots In some ways,
it makes the reading experience feel like a deeper exploration of the human
condition\ldots''

\smallskip
(a) Initially enjoyed podcasts, then maintained interest steadily, now
rekindled.

\textbf{(c) Initially enjoyed podcasts, then lost interest, now rekindled.}

(b) Initially found them uninteresting, then got into them, now rekindled.

(d) No initial mention of enjoying them, then lost interest, now rekindled. &
\textbf{(c)} ``I see you've gone through quite a journey with your interest in
book podcasts. Initially, you enjoyed listening to them, but then there was a
period when you lost interest. Now, it's great to hear that your interest has
been rekindled\ldots'' &
$\checkmark$ \\

\midrule

\textit{Generalize to New Scenarios} &
\textit{Query:} ``I'm considering whether I should start scripting my cooking
videos or just go with the flow. What do you think?''

\smallskip
(a) Scripting provides structure and consistent flow.

(b) Scripting makes editing easier.

\textbf{(c) Going with the flow allows creativity and authenticity.}

(d) A hybrid approach gives the best of both worlds. &
\textbf{(c)} ``Going with the flow might allow for more creativity and
authenticity in your cooking videos\ldots If you enjoy expressing yourself
freely, you might find this approach more fulfilling.'' &
$\checkmark$ \\

\midrule

\textit{Preference-Aligned Recommendation} &
\textit{Query:} ``I'm planning a weekend getaway and want to try something
creatively fulfilling. What would you suggest?''

\smallskip
\textbf{(a) Capturing nature soundscapes in a forest or by a lake.}

(b) Traditional island cooking or Polynesian tattoo design.

(c) Painting in an art studio or backyard.

(d) Writing stories or poetry in a cozy nook. &
\textbf{(a)} ``How about diving into a soundscape adventure by capturing the
symphony of nature\ldots blending the serene rustling of leaves, the melodic
rush of water streams\ldots'' &
$\checkmark$ \\

\midrule

\textit{Suggest New Ideas} \newline{\small(\textit{failure case})} &
\textit{Query:} ``How can I find a more fulfilling way to express my love
for music?''

\smallskip
(a) Music criticism: writing album reviews.

\textbf{(b) Writing about your musical journey or performing live.}

(c) Collaborating with musicians from diverse backgrounds.

(d) Exploring sound engineering and digital remixes. &
\textbf{(c)} ``Collaborating with others who share your musical interests\ldots
A user found fulfillment through working with musicians from diverse
backgrounds\ldots'' \newline\smallskip
{\small\textit{Correct: (b)}} &
$\times$ \\

\bottomrule
\end{tabular}
\end{table*}

\noindent The failure example on \textit{Suggest New Ideas} is insightful: proposing genuinely novel activities requires reasoning about what is \emph{absent} from the user's history rather than what is present. This requires a negation operation in the reasoning process that goes beyond simply surfacing memorized preferences.

\begin{table*}[h]
\centering
\small
\caption{Training hyperparameters and compute requirements for \ourmethod{} on LoCoMo and PersonaMem-v1.}
\label{tab:hyperparameters}
\renewcommand{\arraystretch}{1.15}
\begin{tabular}{lcc}
\toprule
\textbf{Hyperparameter} & \textbf{LoCoMo} & \textbf{PersonaMem-v1} \\
\midrule
Embedding model & Qwen3-Embedding-0.6B & BERT-base-uncased \\
\midrule
Epochs & 8 & 4 \\
Batch size & 1 & 4 \\
Gradient accumulation steps & 4 & 4 \\
Learning rate & $3e^{-3}$ & $3e^{-3}$ \\
Target entropy & $H_{\text{target}}=0.4\,\log k$ & $H_{\text{target}}=0.4\,\log k$ \\
\midrule
Number of memory slots & 32 & 32 \\
Memory dimension & 1024 & 256 \\
Number of memory heads & 64 & 8 \\
Tokens per chunk & 128 & 64 \\
Entropy loss weight & 0.1 & 0.1 \\
\midrule
GPU used & 1$\times$ NVIDIA A40 (40GB) & 1$\times$ NVIDIA A40 (40GB) \\
GPU hours & 7.52 & 2.16 \\
\bottomrule
\end{tabular}
\end{table*}

\begin{table*}[h]
\centering
\small
\setlength{\tabcolsep}{8pt}
\renewcommand{\arraystretch}{1.15}
\caption{Resources used in this work.}
\label{tab:resources}
\begin{tabular}{ll}
\toprule
\textbf{Resource} & \textbf{Link} \\
\midrule
Qwen family of models & https://huggingface.co/Qwen \\
Draw.io & https://www.drawio.com \\
Claude Opus 4.7 & https://www.anthropic.com \\
PersonamemV1 & https://huggingface.co/datasets/bowen-upenn/PersonaMem \\
Locomo dataset & https://github.com/snap-research/locomo \\
\bottomrule
\end{tabular}
\end{table*}
\section{Resources}
The resources used for our work are summarized in Table \ref{tab:resources}.

\end{document}